\relax
\documentclass[letterpaper]{article} 
\usepackage{aaai22}  
\usepackage{times}  
\usepackage{helvet}  
\usepackage{courier}  
\usepackage[hyphens]{url}  
\usepackage{graphicx} 
\usepackage{natbib}  
\usepackage{caption} 
\DeclareCaptionStyle{ruled}{labelfont=normalfont,labelsep=colon,strut=off} 
\usepackage{amssymb}
\usepackage{color}
\usepackage{amsmath}
\usepackage[ruled,vlined]{algorithm2e}
\usepackage{graphicx}
\newtheorem{theorem}{Theorem}
\newtheorem{lemma}[theorem]{Lemma}
\definecolor{commentcolor}{RGB}{75, 104, 104} 
\newcommand{\PyComment}[1]{\ttfamily \textcolor{commentcolor}{\# #1}}  
\newcommand{\PyCode}[1]{\ttfamily \textcolor{black}{#1}} 
\title{Improving Entropy-Based Test-Time Adaptation from a Clustering View}
\author{
    Guoliang Lin,
   Hanjiang Lai,
   Yan Pan,
   Jian Yin
}
\affiliations{
    Sun Yat-sen University\\ lingliang@mail2.sysu.edu.cn \\
    \{  laihanj3, panyan5, issjyin\} @mail.sysu.edu.cn
 \\
}

\usepackage{bibentry}

\begin{document}
\maketitle
\begin{abstract}
    Domain shift is a common problem in the realistic world, where training data and test data follow different data distributions. To deal with this problem, fully test-time adaptation (TTA) leverages the unlabeled data encountered during test time to adapt the model. In particular, entropy-based TTA (EBTTA) methods, which minimize the prediction's entropy on test samples, have shown great success. 
    In this paper, we introduce a new clustering perspective on the EBTTA. 
    It is an iterative algorithm: 1) in the assignment step, the forward process of the EBTTA models is the assignment of labels for these test samples, and 2) in the updating step, the backward process is the update of the model via the assigned samples. This new perspective allows us to explore how entropy minimization influences test-time adaptation.  
    Accordingly, 
    this observation can guide us to put forward the improvement of EBTTA. We propose to improve EBTTA from the assignment step and the updating step, where robust label assignment, similarity-preserving constraint, sample selection,  and gradient accumulation are proposed to explicitly utilize more information. 
    Experimental results demonstrate that our method can achieve consistent improvements on various datasets. Code is provided in the supplementary material. 
\end{abstract}

\section{Introduction}
Deep neural networks (DNNs) have been increasingly applied to various fields in computer vision, including classification~\cite{he2016deep,dosovitskiy2020image}, segmentation~\cite{kirillov2023segment} and detection~\cite{redmon2016you,ren2015faster}. DNNs generalize well when training (source) data and test (target) data follow the same distribution~\cite{he2016deep}. However, the IID assumption might not be held in the real world. For example, test samples may be corrupted by unexpected conditions, like lighting changes or camera moving~\cite{hendrycks2019benchmarking} in practice. This phenomenon
is also referred to as \textit{domain shift} problem, which can easily result in a performance drop.  
Therefore, how to deal with the domain shift problem becomes an urgent demand for DNN deployment in the realistic world. 

One of the research fields that has been put forth is the \textit{ fully test-time adaptation} (TTA), where data from the source domain no longer exists but we can access the source model and unlabeled test data to adapt the model at test time. Various kinds of TTA methods~\cite{wang2021tent,iwasawa2021test,wang2022continual,tomar2023tesla} have been proposed to address the domain shift problem. 
For instance, T3A~\cite{iwasawa2021test} classified  the test sample by computing the distance to the pseudo-prototypes. TeSLA~\cite{tomar2023tesla} was a test-time self-learning method for adapting the source model. Among these methods, entropy-based TTA (EBTTA)~\cite{wang2021tent,niu2022efficient,niu2023towards} methods are  simple and effective approaches. Given test samples, EBTTA only conducts entropy minimization~\cite{niu2023towards} on these test samples. The most representative work of EBTTA is TENT~\cite{wang2021tent}, which modulates the Batch Normalization (BN)~\cite{bjorck2018understanding} layers by minimizing the prediction's entropy using gradient descent. 

Despite the success of EBTTA, these previous studies still leave an open question: why or in which case the entropy minimization~\cite{wang2021tent} can improve the TTA. Simply reducing the entropy may not always succeed. For example, the performance of EBTTA would still vary greatly across different situations.  Zhao et al.~\cite{pitfall} had shown that TENT holds empirical sensitivity and batch dependency through extensive experiments. In Table~\ref{collapse}, it even performs much worse than the source model when the batch size is 1.  Hence, it is necessary to gain a more thorough understanding of EBTTA, and then come up with a better solution.

In this paper, we provide a new perspective to better understand EBTTA through a clustering view\cite{guo2017deep}.  
The entropy-based methods and clustering both belong to unsupervised learning, where entropy minimization tries to reduce uncertainty about which category the sample belongs to, and clustering aims to map data into different clusters. Observing that, we show that entropy minimization and clustering have a lot in common, which makes it possible to use a clustering view to revisit the EBTTA. \textbf{Based on the clustering view, the key components of entropy-based methods are not only at minimizing the entropy loss, but also at implicitly using assigned multiple samples to update the model.} More specifically, we can view the forward process of the deep model in entropy minimization as the assignment step in clustering. According to Lemma~\ref{entropy_lemma}, the entropy loss would increase the probability of the class with the largest value, which is similar to assigning the sample to its nearest clustering center. And the backward process (loss back-propagation and gradient descent) as the center updating step in clustering (see Section~\ref{Interpretation_EB} for detailed discussions). 
The entropy-based methods implicitly use the assigned multiple samples to update the model. This is the key to the success of the entropy-based methods.  

Based on the interpretation, we can set the stage for improving the EBTTA. According to the clustering view, we improve the EBTTA from the assignment step and the updating step. In the assignment step, 
according to the observation in Lemma~\ref{entropy_lemma}, the accurate assignment is very important for EBTTA. In this paper, we propose robust label assignment and similarity-preserving constraint to get a better initial assignment. The data augmentation is first used to obtain a robust initial probability for each test sample. Similarity-preserving constraint can promote that similar samples in feature space are assigned in the same cluster. In the updating step, the key is how to efficiently and stably use the assigned multiple samples to update the model. In this paper, 
we put forward to take advantage of the reliability of samples and the information from more samples, where  sample selection and gradient accumulation are proposed to further improve the EBTTA. 
In summary, we propose to interpret the EBTTA from the view of clustering, which can guide us to develop better approaches for EBTTA. 




Our contributions can be summarized as :

\begin{itemize}
    \item We revisit EBTTA from a clustering view: the forward process can be interpreted as the label assignment while the backward process of EBTTA can be interpreted as the updating of centers in clustering methods.
    \item The clustering view can guide us to put forward the improvement of EBTTA, where the robust label assignment, similarity-preserving constraint, sample selection, and gradient accumulation are proposed to improve EBTTA.
    
    \item We evaluate our method on various benchmark datasets, which demonstrates that our method can gain consistent improvements over existing EBTTA.
\end{itemize}

\section{Related Work}

\subsection{Test-Time Adaptation}

Fully test-time adaptation (TTA) aims to solve the problem of domain shift, where the training (source) domain and the test (target) domain follow different distributions. In order to fulfill this target, TTA online adapts the model using test samples, and a large amount of TTA methods have been proposed. For example, T3A~\cite{iwasawa2021test} and Wang et al.~\cite{wang2023feature} proposed to compute prototypes for each class and online updates the prototypes during adaptations; TeSLA~\cite{tomar2023tesla}, RoTTA~\cite{yuan2023robust} and RMT~\cite{dobler2023robust} enforced the consistency between the teacher and student model; TIPI~\cite{nguyen2023tipi} and MEMO~\cite{zhang2022memo} proposed to learn a model invariant of transformations; SwapPrompt~\cite{ma2024swapprompt} explored the application of TTA on vision-language model. 

The entropy-based test-time adaptation is a simple yet effective practice in  the test-time adaptation, which  minimizes the prediction's entropy of test samples to adapt the model at test time. The most representative work of EBTTA is TENT~\cite{wang2021tent}, which conducted entropy minimizing via stochastic gradient descent and only modulated the BN layers. Then, several methods have been proposed to improve TENT. For example,  ETA~\cite{niu2022efficient} and SAR~\cite{niu2023towards} also minimized prediction's entropy but only selects the reliable samples as inputs. SHOT~\cite{liang2020we} augmented TENT by adding a divergence loss. AdaContrast~\cite{chen2022contrastive} uses nearest neighbor information to generate more correct pseudo labels. CoTTA~\cite{wang2022continual} used TENT to implement continual TTA, where the target domain changes sequentially. Despite the success of EBTTA, several works have found that the performance of EBTTA may sharply drop in some cases, e.g., when adapting on high-entropy samples~\cite{niu2022efficient} or when batch size is small~\cite{wang2021tent,nguyen2023tipi}. To explain this, we propose to view EBTTA as a clustering method and further make improvements on EBTTA based on the clustering view. 

\subsection{Mini-Batch Clustering}
Similar to the TTA, clustering is also a well-known unsupervised method. 
The classifier in TTA and k-means~\cite{bejar2013k} both aim to partition objects into $K$ groups without supervision. When the size of the dataset is large, mini-batch k-means~\cite{bottou1994convergence} also use batches of data to update the centers. Though computationally effective, mini-batch clustering has several well-known disadvantages: (1) Clustering algorithms are highly sensitive to initial centers~\cite{chavan2015mini,celebi2013comparative}. (2) Clustering algorithms are sensitive to outliers since even a few such samples can significantly influence the means of their respective clusters~\cite{celebi2013comparative,yu2011sample}. (3) Mini-batch clustering is also sensitive to batch size, and increasing the batch size can improve the quality of the final partition~\cite{bejar2013k}. 

\subsection{Deep Spectral Clustering}
Spectral clustering~\cite{ng2001spectral,von2007tutorial} is an improvement of k-means algorithm under complex scenarios like the clusters follow a non-Gaussian distribution. Spectral clustering considers the similarity between samples in the same cluster. It uses the \textit{spectral embedding} as data points for subsequent k-means procedure. Compared to the original embedding used for k-means, spectral embedding considers more about samples' nearest neighbor information. 

The spectral clustering is conducted in three steps:

\begin{itemize}
    \item Construct a similarity graph for all samples.
    \item Generate the spectral embeddings of samples using the eigenvectors of Laplacian matrix of the similarity graph.
    \item Apply k-means to partition the spectral embeddings.
\end{itemize}

The classical spectral clustering~\cite{ng2001spectral} stated above needs the calculation for eigenvectors thus lacks scalability. Deep spectral clustering~\cite{shaham2018spectralnet,yang2019deep} is proposed to tackle this problem by learning the spectral embedding with a deep model using the locality-preserving constraint~\cite{huang2014deep} as loss function. This approximation to classical spectral clustering can get better scalability and generalization ability with the cost of accuracy~\cite{shaham2018spectralnet}.

\section{Clustering Interpretation of Entropy-Based TTA}

\subsection{Preliminary}
The classification model $f$ consists of two parts: feature extractor $g$ and classifier head $h$ : $f(x) = h(g(x))$.  We denote the source model as a DNN $\tilde f$ and the target model $f$ is initialized as the source model $f = \tilde f$. The unlabeled target data arrives in the form of batches, and a batch of target samples can be denoted as $X = \{x_1,...,x_N\}$, where $N$ is the batch size. The $P(x) = \delta(f(x)) \in \mathbb{R}^{1 \times K}$ denotes the probabilities of the test sample $x$ belonging to the $K$ classes, where $\delta$ is softmax function. 

Since the ground truth label is unknown, the unsupervised entropy loss $\mathcal{L}_{\mathrm{TENT}}$ for a batch of target samples has been proposed in TENT~\cite{wang2021tent}:
\begin{equation}
  \mathcal{L}_{\mathrm{TENT}}(X) =  \frac{1}{N} \sum_{x \in X}H(x),
    \label{ent}
\end{equation}
where 
\begin{equation}
H(x) = -\sum_{k=1}^{K}[\delta(f(x)]_k \mathrm{log} [\delta(f(x)]_k     
\end{equation}
is the entropy of the $x$. 

\textbf{Mini-Batch K-Means:} We give a brief description of  mini-batch k-means~\cite{bottou1994convergence, bejar2013k}, which is used to interpret EBTTA methods. In the mini-batch k-means, the centers are first initialized, and then the \textbf{A} (assignment) and \textbf{U} (updating) steps are conducted alternately .

Inspired by ~\cite{caron2018deep}, we can theoretically express the learning procedure as:
\begin{gather}
    \min \limits_{C}  \frac{1}{N} \sum_{i=1}^{N} \min \limits_{y_i} \left( \parallel z(x_i) - Cy_i \parallel _2^2 \right) 
    \label{EM}
\end{gather}
where $z(x_i)$ is the feature of sample $x_i$, $y_i$ is the assignment result of $x_i$, and $C$ is the matrix of centers. The problem in \textbf{Eq.~(\ref{EM})} can be solved by an alternating algorithm, fixing one set of variables while solving for the other set. Formally, we can alternate between solving these two sub-problems: 
\begin{equation}
    y_i^j \leftarrow \arg  \min \limits_{y_i}  \parallel z(x_i) - C^{j-1}y_i \parallel _2^2  ,
\label{E}
\end{equation}
\begin{equation}
    C ^j \leftarrow \arg \min \limits_{C} \frac{1}{N} \sum_{i=1}^{N}  \left( \parallel z(x_i) - Cy_i^j \parallel _2^2 \right) ,
\label{M}
\end{equation}
where $j$ is the time step of alternation. We can see that \textbf{Eq.~(\ref{E})} is the \textbf{A-step}: assign the center to each data point by means of the nearest neighbor; and \textbf{Eq.~(\ref{M})} is the \textbf{U-step}: update the centers by averaging features of relevant data points in a batch. 

\begin{figure*}[ht]
    \centering
    \resizebox{\linewidth}{!}{
    \includegraphics{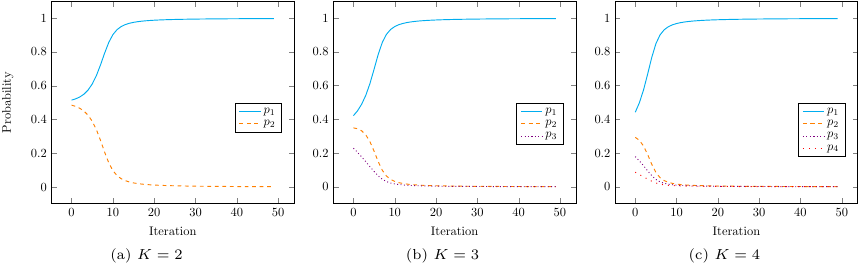}}
    \caption{The probabilities of classes when applying entropy minimization on probabilities via gradient descent, where $K$ is the number of classes. The largest probability will get larger after each iteration of gradient descent.}
    \label{pro}
\end{figure*}

\subsection{Interpretation for Entropy-Based Methods} \label{Interpretation_EB}




We take TENT~\cite{wang2021tent} for example. To easily present our main find, we first consider the binary case, where the number of classes is 2, i.e., $K=2$. Let $c^+$ and $c^-$ denote the positive and negative labels. We denote $p(x;c^+)$ and $p(x;c^-)$ as the probabilities that the test sample $x$ belongs to the positive and negative labels.

\textbf{A-step}: 
In the entropy-based TTA, a test sample goes through the deep model and obtains a $P(x) = \delta(f(x)) = [p(x;c^+), p(x;c^-)]$ in the forward process. This process is similar to the clustering methods that assign the cluster to each data point. EBTTA and clustering methods both assign the nearest label/center via $\hat{c}(x) = \mathop{\arg \max }\limits_{c \in \{c^+,c^-\}}p(x;c)$.  

\textbf{U-step}: With the probability $P(x)$ for each sample $x$, EBTTA uses the entropy loss to update the model. This process is similar to the clustering methods that update the centers. In the clustering methods, the $i$-th center is updated  by averaging features of the samples that were assigned to the $i$-th center in the A-step. We first give a lemma, which shows that the entropy loss also aims to further increase the samples' probability belonging to the $i$-th class, where these samples have the largest probability of the $i$-th class in the forward process. The backward process of EBTTA is very similar to the U-step of clustering, that is both methods use the assigned samples to update the model/centers. 

For the entropy loss, we have the following lemma, which indicates that the entropy loss aims to minimize the uncertainty of each sample that is assigned to the class labels. 
\begin{lemma}
    For sample $x$, the entropy loss $H(x)$ would increase the probability of the class with the largest value, and decrease the sum of probabilities of the other classes .
    \label{entropy_lemma}
\end{lemma}

It is easy to verify in the binary classification.  Without loss of generality, suppose that $p(x;c^{+}) > p(x;c^{-})$ in the A-step. Since SGD is used to update the model, i.e., $p(x;c^{+}) \leftarrow  p(x;c^{+}) - \alpha \frac{dH(x)}{dp(x;c^{+})}$ ($\alpha > 0$ is the learning rate), we only need to show that the gradient of the maximum class is less than zero, i.e., $\frac{dH(x)}{dp(x;c^{+})} < 0$. 

Let $p=p(x;c^{+})$, the derivative of entropy $H(x)$ with respect to $p$ is calculated by:
\begin{align}
    \frac{dH(x)}{dp} &=  \frac{d[ - p \mathrm{log} p - (1-p) \mathrm{log} (1-p) ]}{dp} \\ \notag
    &= \mathrm{log} \frac{1-p}{p}.
\end{align}

According to the assumption $p(x;c^{+}) > p(x;c^{-})$ and $p(x;c^{+}) + p(x;c^{-}) = 1$, we have $p(x_{c^+};c^{+}) > \frac{1}{2}$. Therefore we have $\frac{dH(x)}{dp(x;c^{+})} < 0$ since $\mathrm{log} \frac{1-p}{p} < 0$. The update of the entropy loss by gradient descent can make $p(x;c^{+})$ larger.
%

The proof of multi-classification is given in Appendix A. Lemma~\ref{entropy_lemma} shows that the entropy loss reduces the uncertainty by further increasing the largest probability. For example, if $p(x;c^+) > p(x;c^-)$, the entropy loss will further increase the value of $p(x;c^+)$ and thus decrease the probability of $p(x;c^-)$. Some examples of the changes of probabilities after each iteration via entropy minimization are shown in Figure~\ref{pro}.

\begin{table}[h]
    \resizebox{\linewidth}{!}{
\begin{tabular}{lccc}
\hline
Method & CIFAR-10-C & CIFAR-100-C & ImageNet-C \\ \hline
SOURCE & 56.5       & 8.8         & 18.0       \\
TENT (batch size = 1)   & 10.0       & 1.0         & 0.1        \\ 
TENT (batch size = 20) &78.7 &16.6 &39.5\\
\hline
\end{tabular}
}
\caption{Accuracy(\%) of TENT collapses when batch size is 1.}
\label{collapse}
\end{table}

\textbf{Remark 1}: \textbf{In the A-step, the optimal/minimum of the entropy loss does not mean a good result}. 

As shown in Table ~\ref{collapse}, TENT collapses when batch size is 1 and is worse than SOURCE method even though the entropy loss arrives minimum. 
Lemma 1 shows that if we get only one sample for each iteration (batch size = 1), then the model only tends to make the largest output probability predicted by source model larger until the output probability becomes one-hot as Figure~\ref{pro} shows. More severely, the performance is even worse (will be discussed in Remark 2).  
Thus, what is truly important to the A-step? 

\textbf{Q1: What important information does the A-step provide during adaptation? }

 From the clustering view in the TTA method, the A-step is used to assign the mini-batch samples to the $K$ classes. 
 As we had discussed above, the entropy loss tries to further increase the largest probability of each test sample. Thus the accurate assignment is very important, and we can achieve excellent performance with good assignment, otherwise minimizing the entropy loss may hurt the performances. This is similar to clustering methods that are also sensitive to the initial assignment. 

Several TTA works like AdaContrast~\cite{chen2022contrastive} and TAST~\cite{jang2023testtime} use nearest neighbor information to get more accurate labels. These works are under the assumption that a sample and its nearest neighbor in the feature space are likely to share the same label under the domain shift. The effectiveness of these works can be easily explained from the view of spectral clustering which also considers nearest neighbor information: using spectral clustering can make the clusters robust to complex scenarios~\cite{liu2018spectral}. For example, if data points in a cluster after domain shift follow a non-Gaussian distribution, then it would be a good choice to use spectral clustering to generate clusters.

In summary, from the clustering view, the truly important thing of the A-step is assigning the samples to the cluster centers as accurately as possible. 


\textbf{Remark 2}: \textbf{In the U-step, the good and stable cluster centers are updated via the assigned multiple samples but not single sample}.

When batch size is large than 1, the gradients of multiple samples from target domain will be averaged to update a shared model. Hence, the largest probability of the data point may be changed due to the averaged gradients. This is similar to the clustering method. For example, $x_1$ is assigned to the positive class in the current iteration. After the update of cluster centers, $x_1$ may possibly be assigned to the negative class in the next iteration. This implicit use of multiple samples is an important step for EBTTA. 

\textbf{Q2: How to efficiently and stably use the assigned multiple samples to update the model?}

Several methods, e.g., ETA~\cite{niu2022efficient} and SAR~\cite{niu2023towards}, had proposed to select only the low-entropy samples to efficiently update the model, and the high-entropy samples were removed. The efficiency of this technique has been verified by experiments but a detailed analysis is missing. From the clustering view, we can easily give an explanation: outliers should be suppressed.

The number of the multiple samples is the batch size. The EBTTA methods are also sensitive to the batch size~\cite{wang2021tent,nguyen2023tipi}. This can also be easily answered using the clustering view. As the deep model is updated via the averaged assigned samples' gradients, more samples can provide more stable gradients. That is, when the batch size is small, the variance will be large. This is similar to the clustering methods, where the centers are updated by averaging features of assigned samples. When the number of samples is small, the calculated centers also will have large variance and become unstable. 

 From the view of clustering and the existing methods, using sample selection and larger number of samples become common practices.  

\label{IEBM}
\begin{algorithm}[ht]
\SetAlgoLined
    \PyCode{def forward(x, counter):} \\
    \Indp   
        \PyCode{outputs, features = model(x)} \\
        \PyCode{aug\_outputs, aug\_features = model(aug(x))} \\
        \PyComment{Similarity-Preserving Constraint} \\
        \PyCode{loss\_con = con(features, aug\_features, outputs, aug\_outputs)} \\
        \PyComment{Robust Label Assignment} \\
        \PyCode{outputs = (outputs + aug\_outputs)/2} \\
        \PyCode{ent = softmax\_entropy(outputs)} \\
        \PyCode{loss\_ent = ent} \\
        
        \PyCode{loss = loss\_ent + loss\_con } \\
        \PyCode{mask = (ent <= ent.mean())} \\
        \PyComment{Sample Selection} \\
        \PyCode{loss = loss[mask].mean() } \\
        \PyCode{loss.backward()} \\
        \PyComment{Gradient Accumulation} \\
        \PyCode{if counter \% Q == 0: } \\ 
        \Indp
            \PyCode{optimizer.step()} \\
            \PyCode{optimizer.zero\_grad()} \\
        \Indm
        \PyCode{return outputs} \\
    \Indm 
\caption{Pytorch-like pseudo-code for TTC; $x$ is a batch of unlabeled test samples; $counter$ is the number of batches encountered.}
\label{algo}
\end{algorithm}

\section{Test-Time Clustering (TTC)}

Based on the clustering view and the above discussions, we propose to improve TENT from these two aspects accordingly. In the A-step,  
we suggest to obtain a more robust initial label assignment in order to get high-quality initial probabilities. We also propose to add the similarity-preserving constraint to approximate spectral clustering, which improves the assignment quality. We use these two modules to obtain more accurate assignments. In the U-step, we propose to select test samples to alleviate the negative effects of outliers  Finally, we recommend to use gradient accumulation to overcome the problem of small batch size. 

\subsection{Robust Label Assignment (RLA)}
As discussed in Remark 1, we can see that label assignment in the A-step is very important for the EBTTA, which also provides a good initial probability for each test sample. 

In order to get a better assignment, we simply apply data augmentation inspired by the success of self-supervised learning. Then we combine the two predictions to obtain a more robust probability:
\begin{equation}
    \bar f(x) = \frac{f(x) + f(aug(x))}{2}
\label{ACA}
\end{equation}
where $aug(x)$ is the data augmentation function. Here we only choose horizontal flip which is a weak augmentation function, rather than other strong augmentation functions: 
\begin{equation}
aug(x) = horizontal\_flip(x).    
\end{equation}
 This is because a weakly-augmented unlabeled sample can preserve more information about class label than the strongly augmented functions~\cite{yang2022interpolation}.  

Now the entropy loss for sample $x$ can be calculated as:
\begin{equation}
   \mathcal{L}_{\mathrm{ent}}(x) = \bar H(x) = -\sum_{k=1}^{K}[\delta(\bar f(x)]_k \mathrm{log} [\delta(\bar f(x)]_k  
\end{equation}

\subsection{Similarity-Preserving Constraint (SPC)}
Spectral clustering is an improvement of k-means, which considers the similarity between samples in the same class. It uses the spectral embedding as data points for cluster partitioning. We add the similarity-preserving constraint to generate spectral embedding~\cite{huang2014deep, yang2019deep,shaham2018spectralnet}. Specifically, the constraint loss for a single sample $x$ can be formulated as:
\begin{equation}
    \mathcal{L}_{\mathrm{con}}(x) = \kappa(x,x')\parallel f(x) - f(x')\parallel ^2,
\label{con}
\end{equation}
where $x'$ is the nearest neighbour of $x$ and $\kappa(x,x') = e ^{ - \frac{\parallel g(x) - g(x') \parallel^2}{2 \sigma ^2}}$ is the similarity function, where $g(x)$ is the feature of $x$. We set $\sigma = 1$. By minimizing $\mathcal{L}_{\mathrm{con}}(x)$ can make $f(x)$ be the spectral embedding of sample $x$.

Since spectral clustering especially considers similar samples in the same class, we wish to find the nearest neighbour within the same class. However, if the batch size $N$ is far more less the number of classes $N$ (for example, $N=32$ and $K=1000$), then it is hard to promise the existence of nearest neighbour for each sample $x$ within a batch if we don't store samples from previous batches. Therefore, we use the augmentation version of $x$ as its nearest neighbour as an alternative: $x' = aug(x)$, which can promise that $x$ and $aug(x)$ are in the same class. 

\subsection{Sample Selection (SS)}
Selecting low-entropy samples can not only remove unreliable samples but also improve computational efficiency~\cite{niu2023towards}. However, existing works SAR~\cite{niu2023towards} and ETA~\cite{niu2022efficient} need to pre-define a threshold value and select samples whose entropy is lower than that threshold value. This fixed threshold value needs careful design and may not adapt when the model updates. Instead, we dynamically set the threshold value as the mean entropy of the batch.

To be specific, for a batch of samples $X$, we calculate the mean entropy as:
\begin{equation}
    H_m = \frac{1}{N}\sum_{x \in X} H(x).
\end{equation}

The selecting function $\mathcal{S}(x)$ can be formulated as:
\begin{equation}
    \mathcal{S}(x) = \left\{
    \begin{array}{l}
        1, \ H(x) \leq H_m,  \\
        0, \ H(x) > H_m,
    \end{array}
    \right.
\end{equation}
which means that only samples with entropy no greater than $H_0$ will be selected to update the model.

After applying sample selection, the total loss for our method becomes:
\begin{equation}
    \mathcal{L}_{\mathrm{TTC}}(X) = \frac{1}{N} \sum_{x \in X} \mathcal{S}(x) \cdot [\mathcal{L}_{\mathrm{ent}}(x) + \mathcal{L}_{\mathrm{con}}(x)]
\end{equation}

\subsection{Gradient Accumulation (GA)}
When the batch size is limited, we propose to use gradient accumulation to collect information from multiple batches, which is equivalent to increasing the batch size~\cite{lin2017deep}. In practice, we don't update the parameters immediately once we get the gradient of one batch. Instead, we collect the gradients for $Q$ batches and then conduct gradient descending.

The pseudo-code of our method is shown in Algorithm~\ref{algo}.

\begin{table*}[ht]
\centering
\resizebox{\linewidth}{!}{
\begin{tabular}{lcccccccccccccccc}
\hline
\multicolumn{1}{l|}{ Method } &gaus &shot &impul &defcs &gls &mtn &zm &snw &frst &fg &brt &cnt &els &px &\multicolumn{1}{c|}{jpg} &Avg \\ \hline
\multicolumn{1}{l|}{ SOURCE } &27.7&34.3&27.1&53.0&45.7&65.2&58.0&74.9&58.7&74.0&90.7&53.3&73.4&41.5&\multicolumn{1}{c|}{69.7}&56.5\\
\multicolumn{1}{l|}{ NORM } &71.9&73.8&63.9&87.4&65.0&86.2&87.9&82.7&82.5&84.9&91.9&86.7&76.8&80.6&\multicolumn{1}{c|}{72.9}&79.7\\
\multicolumn{1}{l|}{ T3A } &32.6&38.6&28.9&55.4&48.3&66.5&60.7&74.9&59.6&74.2&90.4&53.9&73.2&42.9&\multicolumn{1}{c|}{69.4}&58.0\\
\multicolumn{1}{l|}{ ETA } &71.8&73.6&63.9&87.2&65.1&86.3&87.7&82.4&82.1&84.8&91.5&87.0&76.2&80.0&\multicolumn{1}{c|}{72.2}&79.5\\
\multicolumn{1}{l|}{ TIPI } &75.2&77.1&67.4&84.7&66.3&82.8&86.1&83.4&82.6&83.5&90.9&85.5&76.8&79.1&\multicolumn{1}{c|}{\textbf{77.8}}&79.9\\ \hline
\multicolumn{1}{l|}{ TENT } &75.7&77.1&67.8&87.7&68.6&86.9&88.6&84.1&83.8&86.2&\textbf{92.0}&\textbf{87.8}&77.6&83.3&\multicolumn{1}{c|}{75.4}&81.5\\
\multicolumn{1}{l|}{ TTC (ours)} &\textbf{77.7}&\textbf{79.4}&\textbf{70.5}&\textbf{88.8}&\textbf{69.5}&\textbf{87.5}&\textbf{89.3}&\textbf{84.3}&\textbf{84.5}&\textbf{87.0}&91.9&87.6&\textbf{79.2}&\textbf{83.8}&\multicolumn{1}{c|}{77.0}&\textbf{82.5}\\ \hline
\end{tabular}
}
\caption{Accuracy(\%) of different methods on CIFAR-10-C with WideResnet28-10.}
\label{c10}
\end{table*}

\begin{table*}[ht]
\centering
\resizebox{\linewidth}{!}{
\begin{tabular}{lcccccccccccccccc}
\hline
\multicolumn{1}{l|}{ Method } &gaus &shot &impul &defcs &gls &mtn &zm &snw &frst &fg &brt &cnt &els &px &\multicolumn{1}{c|}{jpg} &Avg \\ \hline
\multicolumn{1}{l|}{ SOURCE } &16.6&15.0&3.0&8.5&9.9&7.4&8.4&8.5&5.9&1.4&10.8&1.2&9.7&11.7&\multicolumn{1}{c|}{13.6}&8.8\\
\multicolumn{1}{l|}{ NORM } &\textbf{55.1}&\textbf{54.5}&48.5&\textbf{54.2}&\textbf{50.4}&\textbf{52.7}&\textbf{55.5}&\textbf{52.7}&\textbf{54.0}&32.4&\textbf{57.7}&35.7&\textbf{52.2}&\textbf{54.7}&\multicolumn{1}{c|}{\textbf{55.2}}&\textbf{51.0}\\
\multicolumn{1}{l|}{ T3A } &19.5&18.2&5.3&9.7&11.3&9.3&10.5&10.4&7.6&1.9&13.1&1.2&11.7&13.5&\multicolumn{1}{c|}{15.3}&10.6\\
\multicolumn{1}{l|}{ ETA } &50.2&48.1&44.0&49.7&45.8&49.3&50.8&49.1&49.5&43.0&53.1&37.4&48.6&49.1&\multicolumn{1}{c|}{50.2}&47.9\\
\multicolumn{1}{l|}{ TIPI } &38.1&39.7&29.1&43.1&34.2&36.3&38.9&41.6&39.0&13.5&48.6&6.4&38.4&40.9&\multicolumn{1}{c|}{37.8}&35.0\\ \hline
\multicolumn{1}{l|}{ TENT } &52.2&51.8&\textbf{49.3}&50.2&46.2&51.7&52.5&51.7&51.7&\textbf{45.3}&53.7&24.9&47.9&52.8&\multicolumn{1}{c|}{50.2}&48.8\\
\multicolumn{1}{l|}{ TTC (ours)} &53.1&53.2&48.3&52.7&47.7&50.7&53.4&52.1&52.5&45.3&55.3&\textbf{41.1}&50.3&52.0&\multicolumn{1}{c|}{51.6}&50.6\\ \hline
\end{tabular}
}
\caption{Accuracy(\%) of different methods on on CIFAR-100-C with WideResnet28-10.}
\label{c100}
\end{table*}

\begin{table*}[ht]
\centering
\resizebox{\linewidth}{!}{
\begin{tabular}{lcccccccccccccccc}
\hline
\multicolumn{1}{l|}{ Method } &gaus &shot &impul &defcs &gls &mtn &zm &snw &frst &fg &brt &cnt &els &px &\multicolumn{1}{c|}{jpg} &Avg \\ \hline
\multicolumn{1}{l|}{ SOURCE } &2.2&2.9&1.9&17.9&9.8&14.8&22.5&16.9&23.3&24.4&58.9&5.4&17.0&20.6&\multicolumn{1}{c|}{31.6}&18.0\\
\multicolumn{1}{l|}{ NORM } &15.6&16.4&16.3&15.4&15.8&27.1&39.8&35.0&33.8&48.7&65.9&17.3&45.2&50.2&\multicolumn{1}{c|}{40.7}&32.2\\
\multicolumn{1}{l|}{ T3A } &2.2&2.8&1.9&12.8&7.5&9.7&16.6&13.2&17.4&20.4&45.5&4.5&15.4&15.5&\multicolumn{1}{c|}{23.0}&13.9\\
\multicolumn{1}{l|}{ ETA } &25.6&29.9&24.2&4.9&15.7&\textbf{43.2}&49.2&\textbf{49.3}&41.7&58.3&66.0&2.8&\textbf{55.7}&\textbf{59.2}&\multicolumn{1}{c|}{52.7}&38.5\\
\multicolumn{1}{l|}{ TIPI } &17.3&16.9&18.3&13.8&13.9&27.7&42.0&38.3&36.9&51.8&66.6&13.8&49.5&54.5&\multicolumn{1}{c|}{47.9}&34.0\\ \hline
\multicolumn{1}{l|}{ TENT } &27.1&28.8&28.2&26.7&26.0&39.5&48.7&46.2&40.9&57.1&67.5&27.0&53.6&57.9&\multicolumn{1}{c|}{51.5}&41.8\\
\multicolumn{1}{l|}{ TTC (ours)} &\textbf{33.1}&\textbf{33.2}&\textbf{34.7}&\textbf{30.3}&\textbf{29.4}&42.5&\textbf{50.4}&48.0&\textbf{43.4}&\textbf{58.4}&\textbf{68.1}&\textbf{35.2}&54.9&59.2&\multicolumn{1}{c|}{\textbf{53.5}}&\textbf{45.0}\\ \hline
\end{tabular}
}
\caption{Accuracy(\%) of different methods on ImageNet-C with Resnet50.}
\label{ima}
\end{table*}

\section{Experiments}
In this section, we evaluate our method on various datasets. Besides, we have conducted ablation
study to explore the characteristics of our methods. Further, we visualize the feature distribution of different methods. 

\subsection{Datasets and Evaluation Metric}
\subsubsection{Datasets}
We conduct experiments on three common benchmark datasets for TTA evaluation: \begin{itemize}
    \item \textbf{CIFAR-10/100-C}~\cite{hendrycks2019benchmarking}: The source models are trained on the training set of CIFAR-10/100~\cite{krizhevsky2009learning}, which contains 50,000 images with 10/100 classes. Then models are tested on CIFAR-10/100-C, which contains 10,000 images. CIFAR-10/100-C is the corrupted version of the test set of CIFAR-10/100.
    
    \item \textbf{ImageNet-C}~\cite{hendrycks2019benchmarking}: The source models are first trained on the training set of ImageNet~\cite{russakovsky2015imagenet}, which includes 1.2 million images with 1,000 classes. Models are then adapted on the corrupted test set ImageNet-C including 50,000 images. 
\end{itemize}   

Corruption types include Gaussian noise (gaus), shot noise (shot), impulse noise (impul), defocus blur (defcs), glass blur (gls), motion blur (mtn), zoom blur (zm), snow (snw), frost (frst), fog (fg), brightness (brt), contrast (cnt), elastic transform (els), pixelate (px), and JPEG compression (jpg). For all datasets, we choose the highest corruption severity level (level 5) for testing. 

\subsubsection{Evaluation Metric}
We use accuracy to measure the performance of methods on datasets. Here we denote the number
of test samples as $N_{\mathrm{test}}$. Accuracy is a widely used measurement in classification tasks, which can be calculated as:
\begin{equation}
    \mathrm{Accuracy} = \frac{\sum_{i=1}^{N_{\mathrm{test}} } \mathbb{I}_{\{\mathop{\arg \max }P(x_i) == y_i\}}}{N_{\mathrm{test}}},
\end{equation}
where $y_i$ is the true label of $x_i$.
\subsection{Compared Methods}
We compare our method with the following TTA methods: 
\begin{itemize}
    \item SOURCE: the baseline method where no adaptation is conducted.
    \item NORM~\cite{schneider2020improving}(NeurIPS 2020): this method only updates the statistics of BN layers.
    \item T3A~\cite{iwasawa2021test}(NeurIPS 2021): it constructs and updates pseudo-prototype for each class and performs nearest neighbour classification during adaptation.
    \item TENT~\cite{wang2021tent}(ICLR 2021): it updates the BN layers by entropy minimizing via SGD.
    \item ETA~\cite{niu2022efficient}(ICML 2022): it conducts selective entropy minimizing during adaptation.
    \item TIPI~\cite{nguyen2023tipi}(CVPR 2023): this method uses a variance regularizer to adapt the model.
\end{itemize}

\begin{table*}[ht]
\centering
\resizebox{\linewidth}{!}{
\begin{tabular}{lcccccccccccccccc}
\hline
\multicolumn{1}{l|}{ Augmentation } &gaus &shot &impul &defcs &gls &mtn &zm &snw &frst &fg &brt &cnt &els &px &\multicolumn{1}{c|}{jpg} &Avg \\ \hline
\multicolumn{1}{l|}{ RRCrop } &69.1&70.8&61.0&83.3&61.3&81.4&83.3&77.4&76.1&82.1&88.4&79.6&72.1&76.2&\multicolumn{1}{c|}{70.8}&75.5\\
\multicolumn{1}{l|}{ RandRot } &61.7&63.8&52.0&76.3&56.1&75.3&77.8&71.4&71.1&74.6&81.1&73.0&65.7&70.2&\multicolumn{1}{c|}{63.8}&68.9\\
\multicolumn{1}{l|}{ GausNoise } &76.9&79.1&68.6&87.2&\textbf{70.7}&85.2&87.5&\textbf{84.3}&\textbf{84.6}&84.4&90.9&86.2&78.6&\textbf{83.8}&\multicolumn{1}{c|}{\textbf{78.9}}&81.8\\
\multicolumn{1}{l|}{ VerFlip } &62.7&64.7&56.2&77.3&55.5&74.5&78.6&68.9&72.8&75.7&82.5&71.4&65.6&70.4&\multicolumn{1}{c|}{63.2}&69.3\\
\multicolumn{1}{l|}{ HorFlip } &\textbf{77.7}&\textbf{79.4}&\textbf{70.5}&\textbf{88.8}&69.5&\textbf{87.5}&\textbf{89.3}&84.3&84.5&\textbf{87.0}&\textbf{91.9}&\textbf{87.6}&\textbf{79.2}&83.8&\multicolumn{1}{c|}{77.0}&\textbf{82.5}\\ \hline
\end{tabular}
}
\caption{Accuracy(\%) of different methods on CIFAR-10-C with different augmentations.}
\label{tableAugs}
\end{table*}

\begin{table*}[ht]
\centering
\resizebox{\linewidth}{!}{
\begin{tabular}{lcccccccccccccccc}
\hline
\multicolumn{1}{l|}{ Method } &gaus &shot &impul &defcs &gls &mtn &zm &snw &frst &fg &brt &cnt &els &px &\multicolumn{1}{c|}{jpg} &Avg \\ \hline
\multicolumn{1}{l|}{ SOURCE } &70.3&72.5&63.0&\textbf{83.8}&\textbf{80.3}&\textbf{81.0}&\textbf{85.9}&\textbf{83.3}&\textbf{80.2}&42.1&\textbf{88.7}&18.9&\textbf{84.3}&\textbf{87.1}&\multicolumn{1}{c|}{\textbf{88.8}}&74.0\\
\multicolumn{1}{l|}{ NORM } &78.8&78.5&76.5&78.2&73.6&76.6&79.4&75.0&76.0&62.9&81.9&70.4&76.5&78.9&\multicolumn{1}{c|}{79.8}&76.2\\
\multicolumn{1}{l|}{ T3A } &71.3&73.2&64.1&83.3&80.0&80.5&85.3&83.1&80.1&41.0&88.1&19.2&83.7&86.7&\multicolumn{1}{c|}{88.0}&73.8\\
\multicolumn{1}{l|}{ ETA } &78.6&79.6&76.4&79.0&74.7&77.6&79.5&75.1&77.0&64.6&81.0&70.2&76.1&78.4&\multicolumn{1}{c|}{79.6}&76.5\\
\multicolumn{1}{l|}{ TIPI } &73.9&77.7&69.3&67.2&70.2&69.6&72.2&71.4&77.7&50.6&80.6&46.6&71.2&75.8&\multicolumn{1}{c|}{80.1}&70.3\\ \hline
\multicolumn{1}{l|}{ TENT } &77.4&78.2&75.9&75.6&70.8&71.3&80.4&76.5&76.2&72.8&79.9&63.5&72.2&76.9&\multicolumn{1}{c|}{77.8}&75.0\\
\multicolumn{1}{l|}{ TTC (ours) } &\textbf{79.8}&\textbf{80.3}&\textbf{77.6}&79.2&75.5&76.2&80.8&78.8&79.7&\textbf{74.5}&81.9&\textbf{71.2}&76.1&79.8&\multicolumn{1}{c|}{77.6}&\textbf{77.9}\\ \hline
\end{tabular}
}
\caption{Accuracy(\%) of different methods on CIFAR-10-C with Resnet50.}
\label{c10r50}
\end{table*}

\subsection{Implementation Details}
For a fair comparison, all methods use the same model and the same optimizer. Following TIPI~\cite{nguyen2023tipi}, for CIFAR-10-C and CIFAR-100-C, we use model WideResnet28-10~\cite{zagoruyko2016wide} and optimizer Adam~\cite{kingma2014adam} with learning rate 0.001 as a default setting; for ImageNet-C, we use model Resnet50~\cite{he2016deep} and optimizer SGD with learning rate 0.00025 as default. We set the batch size $N = 100$ as default. For TTC, we set $Q = \frac{200}{N}$ and only update BN layers as TENT does. Other details of our method can be found in the provided code.  
\subsection{Comparison on Various Datasets}

The comparison results are shown in Table~\ref{c10}, Table~\ref{c100} and Table~\ref{ima}. The average (avg) accuracy is calculated on all corruption types. The results show that our method can achieve consistent improvements for EBTTA methods w.r.t. average accuracy on all three datasets. More detailed analysis is as follows.

\textbf{CIFAR-10-C}.
The results are shown in Table~\ref{c10}. We can see that TTC achieves the best average accuracy 82.5\%, which is 1.0\% superior to the second-best method TENT. Besides, TTC performs better than TENT on most corruption types.
Specifically, TTC is 2.7\% superior to TENT w.r.t. impulse noise, which is a non-Gaussian transformation. Since TENT is the most relevant work, the comparison results indicate that the proposed  improvements are effective. 

\textbf{CIFAR-100-C}. As shown in Table~\ref{c100}, TTC is 1.8\% better than TENT. It's noting that NORM has achieved a good average accuracy, and our method method has a very close result.

\textbf{ImageNet-C}. The results of Table~\ref{ima} show that TTC achieves the best average accuracy 45.0\%, and the accuracy of the second-best model TENT is 41.8\%. Note that previous entropy-based methods achieve excellent performance, even so, our method performs better than them.


\subsection{Ablation Studies}
In this set of experiments, we conduct ablation studies to analyze the characteristics of our method. 

\subsubsection{Analysis of Different Components}
We conduct experiments to analyze the improvements caused by different components of our method individually. Specifically, we first use TENT as the base model. We apply different components individually to TENT and the average accuracies for datasets are shown in Table~\ref{tableComponents}. We shall see that robust label assignment (RLA) can bring about improvements for TENT on CIFAR-10-C and CIAFR-100-C.  Similarity-preserving constraint (SPC) can improve the performance on CIFAR-100-C and ImageNet-C. Sample selection (SS) can bring about improvements for TENT on both CIFAR-10-C and ImageNet-C.  Gradient accumulation (GA) can make significant improvements on CIFAR-100-C. Please note that the batch size in Table~\ref{tableComponents} is 100, which is not too small. Thus it may be the reason why TENT+GA has similar result to TENT on CIFAR-10-C and ImageNet-C dataset.  Detailed analysis of GA for different batch sizes can be found in Section~\ref{analysisBatchSize}, where GA can boost the performance significantly when the batch size is small.

\begin{table}[h]
\centering
\resizebox{\linewidth}{!}{
\begin{tabular}{lccc}
\hline
\multicolumn{1}{l|}{ Method } & CIFAR-10-C & CIFAR-100-C &ImageNet-C   \\ \hline
\multicolumn{1}{l|}{ TENT  }  &81.5 &48.8 &41.8  \\
\multicolumn{1}{l|}{ \quad +RLA } &82.6({\color{red}+1.1}) &49.2({\color{red}+0.4}) &41.7({\color{green}-0.1}) \\
\multicolumn{1}{l|}{\quad +SPC } &81.5 &49.4({\color{red}+0.6}) &42.2({\color{red}+0.5}) \\
\multicolumn{1}{l|}{\quad +SS} &81.8({\color{red}+0.3}) &47.8({\color{green}-1.0}) &42.8({\color{red}+1.0})   \\
\multicolumn{1}{l|}{\quad +GA} &81.2({\color{green}-0.3}) &51.7({\color{red}+2.9}) &41.6({\color{green}-0.2})   \\ \hline
\end{tabular}}
\caption{Effects of different components applied to TENT individually.}
\label{tableComponents}
\end{table}

\begin{figure*}[ht]
    \centering
    \resizebox{\linewidth}{!}{
    \includegraphics{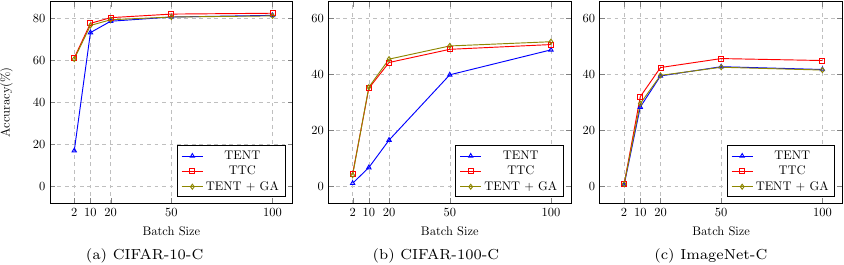}}
    \caption{Accuracy(\%) on various datasets with different batch sizes. Gradient accumulation (GA) can boost performance significantly when the batch size is small. }
    \label{bs}
\end{figure*}

\begin{figure*}[ht]

	\centering

	        \includegraphics[scale=1.15]{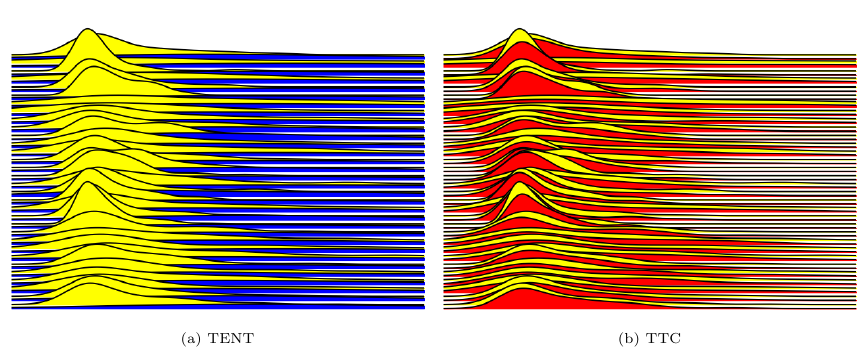}
   \caption{Density plots of test-time feature distribution on  CIFAR-10-C with impulse noise. The color for the feature distribution of the supervised learning as reference is yellow. The color for the feature distribution of TENT is blue. The color for the feature distribution of TTC is red. Each horizontal axis represents a channel. More overlapping areas means more alignment, hence TTC is more aligned with supervised learning.}
\label{density}
\end{figure*}
\subsubsection{Analysis of Different Augmentations}
We conduct experiments to explore the effect of different augmentation functions $aug(x)$. The augmentation functions are horizontal flip (HorFlip), vertical flip (VerFlip), random rotation (RandRot), random resized crop (RRCrop), and Gaussian noise (GausNoise). From the average accuracies for datasets in Table~\ref{tableAugs}, we shall see that horizontal flip makes the performance better than other augmentation functions applied to TTC.

\subsubsection{Analysis of Different Batch Sizes}
We further conduct an ablation study to explore the effect of the batch size, and the average accuracies on datasets are shown in Figure~\ref{bs}. We shall see that all methods perform better when batch size increases. Moreover, when applying gradient accumulation (GA) for small batch sizes, we can get a large performance improvement for TENT. For example, when the batch size is 2, TENT only gets 17.1\% accuracy on CIFAR-10-C. However, when applying GA, we can get an improvement of 43.6\% on TENT.

\label{analysisBatchSize}

\subsubsection{Analysis of Different Model}

We also test our method on CIFAR-10-C using another model ResNet50~\cite{he2016deep}. The results are shown in Table~\ref{c10r50}, which demonstrates that our methods can keep leading performance with various models compared to other methods w.r.t. average accuracy. Specifically, TTC is 2.9\% better than TENT on CIFAR-10-C with Resnet50.

\subsection{Feature Visualization}
We follow ~\cite{wang2021tent,nguyen2023tipi,mirza2022norm} to draw the density plots of the feature distributions. Figure~\ref{density} compares the feature distributions obtained by TENT and TTC on CIFAR-10-C with impulse noise. The feature distribution for supervised learning serves as a reference target. We can see that the feature distribution of our method can be more aligned with that of supervised learning, which implies a better performance. 

\section{Conclusion}
In this paper, we revisited existing EBTTA methods from the view of clustering. We found that the forward process can be interpreted as the label assignment while the backward process of EBTTA can be interpreted as the update of centers in clustering. This view can guide us to improve EBTTA by overcoming weaknesses in clustering methods. We proposed to assign robust labels, use spectral embedding, select samples and accumulate gradients. Conducted experimental results have demonstrated the superiority of our proposed method. In our future work, we wish to give a better theoretical explanation for more test-time adaptation methods.


\bibliography{main}
\clearpage

\appendix
\section{The Proof of Multi-classification in the Lemma 1 }

We assume that in probabilities $P = \{p_i\}_{i=1}^{K}$ (we omit the symbol $x$ for simplicity), the change of the largest one $p_m$ is the opposite of all the other possibilities, for example, when $p_m$ increases, $\{p_i\}_{i \neq m}$ decrease (a natural result of the softmax function):
\begin{gather}
    p_1 \leftarrow p_1 + dp_1 \\ \notag
    \vdots \\ \notag
    p_m \leftarrow p_m + dp_m \\ \notag
    \vdots \\ \notag
    p_K \leftarrow p_K + dp_K ,
\end{gather}
where $dp_m>0$, $dp_i<0$($i \neq m$). 

Since the probabilities sum up to 1, we have:
\begin{equation}
    p_m + \sum_{i \neq m} p_i  = 1, 
\end{equation}
and 
\begin{equation}
    p_m + dp_m + \sum_{i \neq m} (p_i + dp_i)  = 1.
\end{equation}
Therefore, 
\begin{equation}
   - \sum_{i \neq m}\frac{dp_i}{dp_m} = - \frac{d\sum_{i \neq m}p_i}{dp_m} = - \frac{d(1-p_m)}{p_m} = 1.
\label{eq1}
\end{equation}

The total derivative of $H$ w.r.t. $p_m$ is:
\begin{align}
    \frac{dH}{dp_m}(P)= & \sum_{i=1}^{K}\frac{\partial H}{\partial p_i}(P) \cdot \frac{dp_i}{dp_m} \\ \notag
    =& \frac{\partial H}{\partial p_m}(P) + \sum_{i \neq m}\frac{\partial H}{\partial p_i}(P) \cdot \frac{dp_i}{dp_m} \\ \notag
    =& -(1 + \mathrm{log}(p_m)) + \sum_{i \neq m} \frac{dp_i}{dp_m}[-(1 + \mathrm{log}(p_i)] \\ \notag
    =& -(1 + \mathrm{log}(p_m)) -\sum_{i \neq m} \frac{dp_i}{dp_m} \\ \notag
    &- \sum_{i \neq m} \frac{dp_i}{dp_m}\mathrm{log}(p_i). \\
\label{eq2}
\end{align}
Substituting \textbf{Eq.~(\ref{eq1})} into \textbf{Eq.~(\ref{eq2})} we have:
\begin{equation}
    \frac{dH}{dp_m}(P) = -\mathrm{log}(p_m) - \sum_{i \neq m} \frac{dp_i}{dp_m}\mathrm{log}(p_i).
\label{eq3}
\end{equation}

Suppose that the second largest probability $p_n = \mathrm{max}(\{p_i\}_{i \neq m})$, then considering $\frac{dp_i}{dp_m}<0(i \neq m)$ and \textbf{Eq.~(\ref{eq1})} we have: 
\begin{equation}
    - \sum_{i \neq m} \frac{dp_i}{dp_m}\mathrm{log}(p_i) \leq - \sum_{i \neq m} \frac{dp_i}{dp_m}\mathrm{log}(p_n) = \mathrm{log}(p_n).
\label{eq4}
\end{equation}

Since the largest probability $p_m$ must hold $p_m > p_n$, substituting \textbf{Eq.~(\ref{eq4})} into \textbf{Eq.~(\ref{eq3})} we have :
\begin{align}
    \frac{dH}{dp_m}(P) \leq -\mathrm{log}(p_m) + \mathrm{log}(p_n) < 0.
\end{align}
Therefore, the minimizing of entropy loss via gradient descent can increase the largest probability.
\end{document}